\documentclass{article} 
\usepackage{iclr2027_conference,times}
\iclrfinalcopy

\usepackage{amsmath,amsfonts,bm}

\def\eqref#1{equation~\ref{#1}}

\def\1{\bm{1}}

\DeclareMathAlphabet{\mathsfit}{\encodingdefault}{\sfdefault}{m}{sl}
\SetMathAlphabet{\mathsfit}{bold}{\encodingdefault}{\sfdefault}{bx}{n}

\definecolor{citecolor}{HTML}{0071bc}
\usepackage[hyphens]{url}
\usepackage[breaklinks=true,colorlinks,citecolor=citecolor]{hyperref}  

\usepackage{graphicx}
\usepackage{fontawesome5}
\usepackage[caption=false,font=small]{subfig}
\usepackage{wrapfig}
\usepackage{hyperref}
\usepackage{url}
\usepackage{lipsum}

\newcommand{\resourceicon}[1]{\makebox[1.2em][c]{#1}\hspace{0.3em}}
\newcommand{\passat}[1]{\ensuremath{\mathrm{pass@}#1}}

\usepackage{array,booktabs}
\usepackage{listings}
\usepackage{algorithm}
\usepackage{algpseudocode}
\usepackage[breakable,skins]{tcolorbox}



\definecolor{PaperBoxCoreRed}{HTML}{E21500}
\definecolor{PaperBoxDarkRed}{HTML}{5A0800}
\definecolor{PaperBoxGray200}{HTML}{D4D4D4}
\definecolor{PaperBoxGray600}{HTML}{5A5A5A}
\definecolor{PaperBoxGray800}{HTML}{2D2D2D}
\definecolor{PaperBoxBlack}{HTML}{000000}
\definecolor{PaperBoxWhite}{HTML}{FFFFFF}

\tcbset{paper statement/.style={
  enhanced,
  breakable,
  boxrule=0.5pt,
  arc=1mm,
  auto outer arc,
  coltext=PaperBoxBlack,
  coltitle=PaperBoxWhite,
  fontupper=\normalfont,
  fonttitle=\normalfont\bfseries,
  boxsep=1mm,
  left=1mm,
  right=1mm,
  top=1mm,
  bottom=1mm,
  toptitle=0pt,
  bottomtitle=0pt,
  titlerule=0pt,
  before skip=7pt plus 1pt minus 1pt,
  after skip=7pt plus 1pt minus 1pt,
  parbox=false,
}}

\newcommand{\paperboxtitle}[2]{%
  #1\if\relax\detokenize{#2}\relax\else~#2\fi}

\newtcolorbox{observationbox}[1][]{
  paper statement,
  colframe=PaperBoxGray600,
  colbacktitle=PaperBoxGray600,
  colback=PaperBoxGray200!18!PaperBoxWhite,
  title={\paperboxtitle{Observation}{#1}},
  title after break={\paperboxtitle{Observation}{#1} (continued)},
}

\newtcolorbox{takeawaybox}[1][]{
  paper statement,
  colframe=PaperBoxGray800,
  colbacktitle=PaperBoxGray800,
  colback=PaperBoxGray200!10!PaperBoxWhite,
  title={\paperboxtitle{Takeaway}{#1}},
  title after break={\paperboxtitle{Takeaway}{#1} (continued)},
}

\newtcolorbox{discussionbox}[1][]{
  paper statement,
  colframe=PaperBoxDarkRed,
  colbacktitle=PaperBoxDarkRed,
  colback=PaperBoxDarkRed!4!PaperBoxWhite,
  title={\paperboxtitle{Discussion}{#1}},
  title after break={\paperboxtitle{Discussion}{#1} (continued)},
}

\newtcolorbox{guidancebox}[1][]{
  paper statement,
  colframe=PaperBoxGray800,
  colbacktitle=PaperBoxGray800,
  colback=PaperBoxGray200!10!PaperBoxWhite,
  title={\paperboxtitle{Practical Guidance}{#1}},
  title after break={\paperboxtitle{Practical Guidance}{#1} (continued)},
}

\title{Elicitation or Expansion? Demystifying Gains from Agent Harness Optimization\texorpdfstring{\vspace{3pt}}{}}
\author{%
\parbox{\dimexpr\textwidth-2\tabcolsep\relax}
{\raggedright\normalfont
{\bfseries
Younghyun Kim\textsuperscript{1}\thanks{Equal contribution.\quad
\ensuremath{^{\dagger}}Co-corresponding authors.}\hspace{0.4em},
Wooseok Seo\textsuperscript{1, 2}\ensuremath{^{*}},
Minseo Kim\textsuperscript{3},
Seunghyuk Oh\textsuperscript{1},
Minjae Lee\textsuperscript{1}\\[2pt]
Donghoon Kim\textsuperscript{1},
Hyung Il Koo\textsuperscript{1},
Jaehyung Kim\textsuperscript{2}\ensuremath{^{\dagger}},
Wonjun Kang\textsuperscript{1}\ensuremath{^{\dagger}}}\\[5pt]
{\small\textsuperscript{1}FuriosaAI\quad
\textsuperscript{2}Yonsei University\quad
\textsuperscript{3}University of California, Berkeley}\\[3pt]
{\small
\resourceicon{\textcolor{citecolor}{\faGlobe}}Project page: \href{https://furiosa-air.github.io/elicitation-or-expansion}{\textcolor{citecolor}{\nolinkurl{furiosa-air.github.io/elicitation-or-expansion}}}\\
\resourceicon{\faGithub}Code: \href{https://github.com/furiosa-ai/elicitation-or-expansion}{\textcolor{citecolor}{\nolinkurl{github.com/furiosa-ai/elicitation-or-expansion}}}\\
\resourceicon{\raisebox{-0.15em}{\includegraphics[height=1em]{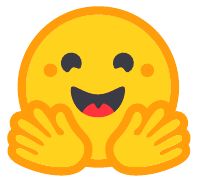}}}Dataset: \href{https://huggingface.co/datasets/furiosa-ai/elicitation-or-expansion-trajectories}{\textcolor{citecolor}{\nolinkurl{datasets/furiosa-ai/elicitation-or-expansion-trajectories}}}\endgraf}%
}}
\hypersetup{
pdftitle={Elicitation or Expansion? Demystifying Gains from Agent Harness Optimization},
pdfauthor={Younghyun Kim, Wooseok Seo, Minseo Kim, Seunghyuk Oh, Minjae Lee, Donghoon Kim, Hyung Il Koo, Jaehyung Kim, Wonjun Kang}
}

\begin{document}

\maketitle
\fancyhead{}
\lhead{Preprint}
\renewcommand{\headrulewidth}{0.4pt}

\vspace{-0.2in}
\begin{abstract}
\vspace{-0.1in}
Recent studies in harness optimization have shown substantial performance gains in language model agents without updating model weights.
A well-designed harness more effectively leverages the underlying model's capabilities by shaping context, tool use, and execution workflows.
However, na\"ive success rates alone do not fully characterize whether optimized harness yields more reliable success on tasks already within reach (\emph{elicitation}) or a broader range of solvable tasks (\emph{expansion}).
This distinction is crucial for recursive self-improvement and evolutionary search, as expansion may provide a broader range of successful experiences to push the capability boundaries.
Despite the growing significance, this distinction remains underexplored in harness optimization.
We conduct a systematic analysis to study the effect of gains from harness optimization.
Building on \passat{k} analyses in the reinforcement learning (RL) literature that compare models before and after RL training, we compare optimized harnesses with their pre-optimization counterparts.
Specifically, we assess changes in reachable task coverage, compare harness optimization with model scaling and increased inference-time compute, and examine how gains vary with task difficulty.
Across static prediction and dynamic tasks, we find that \passat{1} advantages of harness over stronger interventions can reverse at larger attempt budgets and aggregate gains conceal shortfalls on hard tasks, underscoring the need for a fine-grained diagnosis.
\end{abstract}

\vspace{-0.1in}
\section{Introduction}
\label{sec:introduction}

The ability of a language model agent is influenced by the design of its \emph{harness}, which governs prompting, context management, tool use, and execution logic~\citep{young2025harness,openai2026harnessengineering}.
Refining these components can substantially improve agent performance without updating model parameters, motivating growing interest in harness optimization.
By broadening the scope of harness optimization beyond prompts~\citep{agrawal2026gepa} and context~\citep{zhang2026agentic} to agent code and execution workflows~\citep{hu2025automated,lee2026metaharness}, recent methods have achieved huge performance gains across diverse tasks.
However, these aggregate benchmark gains alone do not reveal which aspects of the agent system have improved.
For instance, comparable gains arise either from simply reformatting otherwise correct answers to satisfy the evaluation protocol or from enabling the agent to more effectively leverage its environment and external feedback to deal with previously unsolved tasks.
This distinction is especially important for approaches that leverage successful experience for subsequent improvement, including recursive self-improvement ~\citep{zhang2026darwin,lee2026recursive}.
In these settings, even rare successes on previously unreachable tasks can provide a foundation for further search or learning.

Recent studies have begun to examine the nature of harness gains by relating prompt edits to task-dependent performance changes~\citep{gong2026prompt}, attributing improvements to recoverable execution defects~\citep{yang2026one}, and comparing harness evolution with simple retries~\citep{wang2026rethinking}.
However, these analyses leave open how task coverage under the original and optimized harnesses differs as the attempt budget increases.
Whether the task coverage of optimized harness persists, disappears, or reverses matters for assessing how far the optimized harness can extend an agent's practical problem-solving reach without further model training.
A closely related debate in reinforcement learning (RL) provides a useful diagnostic perspective.
Comparisons between RL-trained models and their pre-RL counterparts show that improvements in \passat{1} can diminish or reverse at larger sampling budgets, suggesting improved sampling efficiency rather than broader task coverage~\citep{yue2025does}.
Other studies report broader coverage under the valid reasoning evaluation~\citep{wen2026reinforcement} or additional tool use~\citep{zhai2026does}, while controlled experiments show that coverage gains depend on pre-training exposure and training-task difficulty~\citep{zhang2026on}.
Together, these studies lay the ground for examining how improvements behave across sampling budgets and task conditions, rather than inferring their nature from a single aggregate score.

\begin{wrapfigure}{r}{0.5\textwidth}
\vspace{-0.2in}
\centering
\input{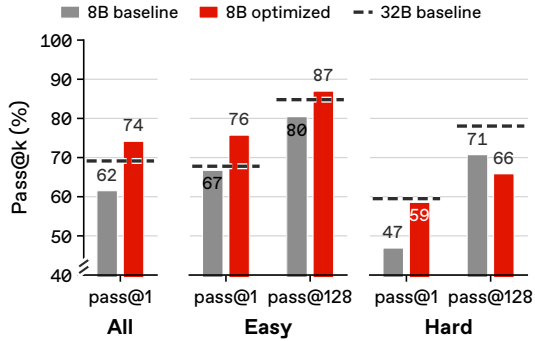}
\vspace{-0.25in}
\caption{On HotpotQA, the optimized harness enables Qwen3-8B to outperform even 32B under base harness in overall \passat{1}.
On easy questions, this advantage holds at both \passat{1} and \passat{128}.
On hard questions, however, the optimized harness outperforms the original 8B baseline at \passat{1} but falls below it at \passat{128}.}
\label{fig:teaser-hotpotqa}
\vspace{-0.1in}
\end{wrapfigure}

In this work, we investigate whether harness optimization makes success more frequent on tasks already solved by the baseline (\emph{elicitation}) or broadens observed task coverage (\emph{expansion}) through the lens of \passat{k}~\citep{chen2021evaluating}.
We complement this analysis by comparing optimized harnesses with base harness using larger models from the same family or higher reasoning effort.
These baselines provide empirical reference for assessing how far harness optimization alone enhances an agent's capability.
Finally, we examine whether gains are concentrated on easier tasks or extend to harder ones. 
Figure~\ref{fig:teaser-hotpotqa} depicts why these multi-axis analyses matter.
Here, overall \passat{1} gain does not fully characterize harness gains; the optimized harness leaves the agent less capable of solving harder problems than under the original harness.
Our analysis thus unveils concealed coverage loss on harder tasks.\footnote{This example uses an optimized harness under relaxed per-attempt resource limits in Appendix~\ref{app:optimization-conditions}}

The heterogeneous effects recur throughout our analyses on various benchmarks.
First, advantages over stronger references can reverse as the attempt budget increases: on LawBench~\citep{fei2024lawbench}, the optimized GPT-OSS-20B harness exceeds the GPT-OSS-120B baseline by $16.05$ percentage points (pp) at \passat{1}, but trails it by $4.00$ pp at \passat{64}.
Second, on HotpotQA~\citep{yang2018hotpotqa} and AppWorld~\citep{trivedi2024appworld}, the selected harnesses approach or exceed stronger references in aggregate but do not surpass them on hard subsets.
Third, the transferred text-classification harness falls below its baseline at \passat{64} on all out-of-distribution datasets, despite small \passat{1} gains on two.
This heterogeneity makes systematic analysis necessary: understanding harness gains requires distinguishing reliability from coverage and examining how both change across attempt budgets, task groups, reference systems.
\vspace{-0.05in}
\section{Related Work}
\label{sec:related-work}

\vspace{-0.05in}
\subsection{Harness Optimization}
Harness optimization adapts agent's prompts, context, and execution logic through repeated evaluation and revision.
TextGrad~\citep{yuksekgonul2025optimizing} and GEPA~\citep{agrawal2026gepa} refine prompts through natural-language feedback, while ACE~\citep{zhang2026agentic} incorporates lessons from execution into an evolving context playbook.
Methods that search over agent code~\citep{hu2025automated,zhang2026darwin,zhang2026darwinx} use execution feedback to diagnose failures and guide revisions across harness components~\citep{lin2026ahe,zhang2026harnesscompass,park2026autosaddler,yue2026ecdysis}.
Our study focuses on Meta-Harness~\citep{lee2026metaharness}, which uses a coding-agent proposer to selectively inspects prior harness code, evaluation scores, and execution traces for new candidates.

\vspace{-0.05in}
\subsection{Interpreting Capability Gains}
Analyses of reinforcement learning show that higher \passat{1} can accompany reduced coverage at large sampling budgets~\citep{yue2025does}, while other studies find broader coverage under different training and evaluation conditions~\citep{wen2026reinforcement,zhang2026on}.
Recent work further documents smaller gains on harder problems under standard reinforcement learning and shows that adaptive sampling can improve learning on these problems~\citep{noukhovitch2026learning}.
Studies of repeated sampling~\citep{brown2024large} and agent interaction budgets~\citep{zhai2026does} examine how additional attempts and interactions affect coverage, while test-time scaling analyses show that effective compute allocation depends on task difficulty~\citep{snell2025scaling}.
For harness optimization, analyses compare evolution with test-time search~\citep{wang2026rethinking}, trace gains to execution repairs~\citep{yang2026one}, and examine the repeatability~\citep{zhao2026promptsmeasuringoptimizingllm} and model dependence~\citep{lin2026harness} of improvements.
We analyze optimized harnesses across attempt budgets and difficulty levels, distinguishing coverage gains over the original baseline from advantages that persist against baselines with larger models or more inference compute.
\section{Analysis Setup}
\label{sec:analysis-setup}

\subsection{Preliminary: Harness Optimization}
\label{sec:harness-optimization}

Let $M$ denote a solver model and $H$ its harness, which controls model inputs, tool interactions, and execution state.
On a task $x$, executing $(\mathcal{M},H)$ yields a complete trajectory $\tau \sim \pi_{\mathcal{M},H}(\cdot \mid x)$, evaluated by a task-specific scoring function $r(\tau,x)$.
Given a task distribution $\mathcal{X}$, harness optimization seeks to maximize the expected task score $J(\mathcal{M},H)$ by revising $H$ while keeping $\mathcal{M}$ fixed:
\[
\max_{H}\; J(\mathcal{M},H), \qquad
J(\mathcal{M},H)
=
\mathbb{E}_{x \sim \mathcal{X},\,
\tau \sim \pi_{\mathcal{M},H}(\cdot \mid x)}
\bigl[r(\tau,x)\bigr].
\]

For our main experiments, we use Meta-Harness~\citep{lee2026metaharness}, whose coding agent, called proposer, selectively inspects previous harness code, evaluation scores, and execution traces to diagnose failures and propose revised harnesses.
Each candidate is evaluated, and its code and execution feedback are retained to guide subsequent proposals.
Our analysis compares the initial harness $H_0$ with an optimized harness $H^\star$ selected based on validation performance.
We estimate their \passat{k} from repeated executions on a common test set under the same task-specific scorer (Appendix~\ref{app:passk}).

\subsection{Motivation: Beyond Mean Accuracy} 
\label{sec:motivation}

\begin{wrapfigure}[15]{r}{0.42\textwidth}
\vspace{-0.42in}
\centering
\includegraphics[width=\linewidth]{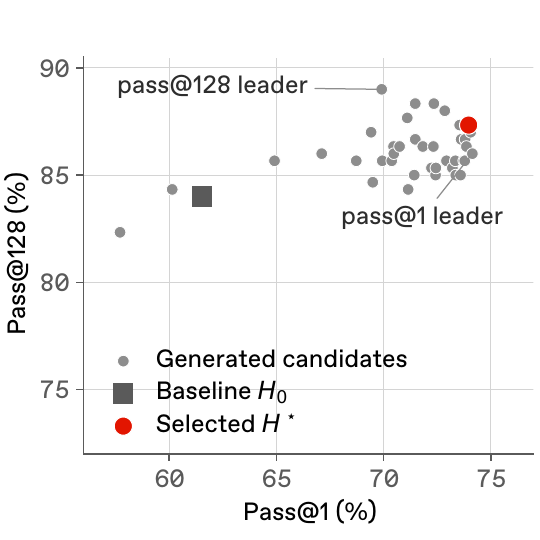}
\vspace{-0.4in}
\caption{Mean accuracy and coverage for harnesses from one optimization run.}
\label{fig:motivation-coverage}
\vspace{-0.2in}
\end{wrapfigure}

For a fixed set of tasks and number of attempts ($k$) per task, \passat{1} determines the expected number of successful executions (\textit{i.e.}, mean accuracy), whereas \passat{k} measures the expected fraction of tasks solved at least once in $k$ complete executions~\citep{chen2021evaluating} (\textit{i.e.}, coverage).
A harness can therefore achieve higher mean accuracy but low coverage, by repeatedly solving the same tasks without obtaining successes on additional ones.
When generated solutions are used for self-training~\citep{zelikman2022star} or verifier-based selection~\citep{cobbe2021verifiers}, coverage is particularly important as it captures the breadth of tasks for which a successful example or solution is available.

Figure~\ref{fig:motivation-coverage} compares \passat{1} and \passat{128} for Qwen3-8B harness candidates from one optimization run, each evaluated on the same 300 HotpotQA tasks with $k=128$ executions per task.
Although the two metrics broadly track one another, the candidate with the highest estimated \passat{1} differs from the one with the highest observed \passat{128}.
Repeated evaluation can improve the precision of \passat{1}, but does not make it a substitute for measuring task coverage at larger budgets.

\section{Evaluation Setup}
\label{sec:evaluation-setup}

We refer to $(\mathcal{M},H_0)$ and $(\mathcal{M},H^\star)$ as the baseline and optimized system, respectively.
The reference $(\mathcal{M}_{\mathrm{ref}},H_0)$ uses the base harness with a larger solver model or the same solver model at higher reasoning effort.
We compare each system on a large $k$ attempts.
More details are in Appendix~\ref{app:protocols}.

\subsection{Static Tasks}
\label{sec:setup-static}

\textbf{Text classification.}
Following \citet{lee2026metaharness}, we jointly optimize a single harness across three static prediction datasets: Symptom2Disease~\citep{gretel2023symptom} for diagnosis from symptom descriptions, LawBench~\citep{fei2024lawbench} for charge-set prediction, and USPTO~\citep{schneider2016whats} for reactant-SMILES set prediction.
In these tasks, the harness constructs reference context to support prediction.
We use a few-shot harness as $H_0$, since directly supplying labeled examples provides a simple way to construct such context.
The optimized harness $H^\star$ prepares memory from the same pool and retrieves relevant information to construct context for each test input.
Across repetitions, $H_0$ resamples demonstrations and $H^\star$ rebuilds memory (Appendix~\ref{app:frozen-memory}).
For out-of-distribution (OOD) transfer, we reuse $H^\star$ obtained from in-distribution (ID) datasets to FiNER~\citep{loukas2022finer}, GoEmotions~\citep{demszky2020goemotions}, and Banking77~\citep{casanueva2020efficient}.
We use GPT-OSS-20B~\citep{agarwal2025gpt} as $\mathcal{M}$ and GPT-OSS-120B as $\mathcal{M}_{\mathrm{ref}}$ for comparison.

\subsection{Dynamic Tasks}
\label{sec:setup-dynamic}

\textbf{Multi-hop QA.} 
On HotpotQA~\citep{yang2018hotpotqa}, the harness coordinates evidence retrieval and answer generation.
The base harness $H_0$ follows multi-hop program of GEPA~\citep{agrawal2026gepa}: it retrieves passages, generates a follow-up query from the retrieved evidence, and answers using summaries accumulated across the two retrieval rounds.
We obtain $H^\star$ by optimizing the harness code governing query construction, evidence processing, and answer generation, while keeping the BM25 retriever fixed.
Evaluation uses 300 test questions with their original difficulty labels.
We use Qwen3-8B~\citep{yang2025qwen3} as $\mathcal{M}$ and Qwen3-32B as $\mathcal{M}_{\mathrm{ref}}$ for comparison.

\textbf{Agentic task.}
On AppWorld~\citep{trivedi2024appworld}, the harness coordinates multi-turn interactions with simulated applications, where actions change the environment and execution feedback informs subsequent decisions.
The baseline harness $H_0$ follows the benchmark's simplified ReAct implementation~\citep{yao2023react}, alternating Python code generation and execution.
Evaluation uses the 168 tasks in \texttt{test\_normal} with their original difficulty labels.
We use GPT-OSS-120B for both $\mathcal{M}$ and $\mathcal{M}_{\mathrm{ref}}$.
The solver's reasoning effort is fixed at medium during harness optimization and for the base and optimized systems; the reference system uses high reasoning effort.
\section{Experimental Results}
\label{sec:experimental-results}

\begin{figure}
\centering
\input{figures/latex/task_comparison.tex}
\caption{
\passat{k} curves for the base solver with baseline ($H_0$) and selected optimized ($H^\star$) harnesses, compared with a stronger reference using $H_0$.
``base'' and ``reference'' denote GPT-OSS-20B and GPT-OSS-120B for text classification (a-c), Qwen3-8B and Qwen3-32B for HotpotQA (d), and GPT-OSS-120B at medium and high reasoning effort for AppWorld (e), respectively.
Each red curve represents a harness selected from one optimization run.
}
\label{fig:task-comparison}
\par
\input{figures/latex/text_classification_difficulty.tex}
\caption{
Distributions of task-level observed success fractions for GPT-OSS-20B with baseline ($H_0$) and optimized ($H^\star$) harnesses.
Bars show the percentage of tasks within each dataset falling in each bin, computed separately for each harness.
}
\label{fig:static-task-difficulty}

\end{figure}

\subsection{Do Optimized Harnesses Provide Gains Over Stronger References?}
\label{sec:results-reference}

\begin{observationbox}
The optimized GPT-OSS-20B harness falls below the GPT-OSS-120B reference at \passat{64} on all three static benchmarks, despite its varying relative performance at \passat{1}.
As shown in Figure~\ref{fig:task-comparison} (a-c), \passat{1} of optimized harness nearly matches the 120B reference on Symptom2Disease, exceeds it on LawBench, and falls below it on USPTO.
LawBench provides a particularly clear example: the $16.05$ pp \passat{1} advantage reverses at $k=11$; at \passat{64}, the optimized harness reaches $56\%$ versus the reference's $60\%$.
Thus, a substantial single-attempt advantage does not necessarily translate into broader coverage under repeated attempts. 
Additional results on static tasks are in Appendix~\ref{app:static-numbers}.

On the dynamic benchmarks, the selected harnesses attain overall \passat{1} and \passat{128} close to those of their stronger references.
On HotpotQA, both optimized Qwen3-8B harnesses exceed the 32B reference at \passat{1}, but only one remains above it at \passat{128}.
On AppWorld, both optimized medium-effort harnesses fall below the high-effort reference at these endpoints, although one briefly exceeds it at $k=2$.
Comparisons within a single optimization run in Figure~\ref{fig:hotpot-evolution}
(a) further illustrate that advantages in single-attempt success and finite-budget coverage do not
coincide. 
Additional results on dynamic tasks are in Appendix
~\ref{app:dynamic-results}.
\end{observationbox}

\begin{discussionbox}

\textbf{Baseline success patterns help explain the heterogeneous gains from harness optimization.}
To interpret the text-classification results, we compare the task-level success distributions of the baseline and optimized harnesses with GPT-OSS-20B in Figure~\ref{fig:static-task-difficulty}.
For each task $x$, we compute its observed success fraction $c/n$ separately under each harness, where $c=\sum_{j=1}^{n}r(\tau_j,x)$.
We then construct a separate distribution for each harness by grouping tasks according to these fractions, distinguishing no observed success, intermittent success, and success on every attempt.
On Symptom2Disease, frequent baseline success leaves limited \passat{1} headroom (Figure~\ref{fig:static-task-difficulty} (a)).
Optimization improves reliable successes but with the slight decrease in \passat{64} coverage.
On LawBench, optimization primarily makes existing successes more frequent (Figure~\ref{fig:static-task-difficulty} (b)).
Of the $53$ tasks solved by the baseline within 64 attempts, the optimized harness retains $49$ and loses $4$, while newly solving $7$ others; the retained tasks account for $21.02$ pp of the $25.14$ pp \passat{1} gain (Appendix~\ref{app:success-decomposition}).
Consequently, the gain over the baseline narrows to $3.00$ pp at \passat{64} (Figure~\ref{fig:task-comparison} (b)).
On USPTO, observed coverage increases at \passat{64}, but $68$ of $100$ tasks still yield no success, leaving substantial headroom unrealized (Figure~\ref{fig:static-task-difficulty} (c)).
These patterns suggest that opportunities for harness improvement depend on both the frequency and breadth of baseline successes.
These patterns motivate a possible analogy to the Matthew effect reported by \citet{noukhovitch2026learning}, where RL gains tend to favor tasks with greater initial model ability.
Harness optimization similarly concentrates gains on tasks with existing baseline successes.

\textbf{Environment interaction provides additional opportunities for harness optimization.}
On dynamic tasks, optimized harness also shapes actions and hence subsequent observations, creating opportunities to acquire additional information.
For instance, in our HotpotQA experiment, the base harness restates the question as its second retrieval query, leaving the ``George Litto'' abstract unreachable in all 128 rollouts and ultimately reporting that the producer of \textit{Blow Out} is not mentioned.
By instead querying the newly resolved bridge entity, the optimized harness reaches the relevant abstract in 125 of 128 rollouts and answers correctly in 122 times, turning previously inaccessible external evidence into usable context.
This aligns with prior work showing that structuring action loops and agent interfaces can improve performance without updating model weights~\citep{yao2023react,yang2024sweagent}, while complementary evidence from tool-use RL suggests that improved sequential information gathering can broaden observed task coverage~\citep{zhai2026does}.
Such opportunities are absent or more limited in the static tasks in Figures~\ref{fig:task-comparison} (a)-(c), where harness optimization mainly reshapes how the solver uses a fixed input.
This contrast may help explain why, in the dynamic tasks in Figures~\ref{fig:task-comparison} (d)-(e), the selected optimized harnesses produce \passat{k} curves that closely track those of the stronger references.
\end{discussionbox}

\begin{takeawaybox}
Harness optimization can yield substantial \passat{1} gains without matching stronger references at larger $k$, especially on static tasks.
On dynamic tasks, improved environment interaction can expand accessible evidence, helping optimized harnesses more closely track stronger references.
\end{takeawaybox}

\subsection{Do Optimized Harness Gains Differ by Task Difficulty?}
\label{sec:results-difficulty}

\begin{figure}
\centering
\input{figures/latex/hotpotqa.tex}
\caption{
HotpotQA \passat{k} curves for the Qwen3-8B baseline ($H_0$) and selected optimized harnesses ($H^\star$), with a Qwen3-32B baseline. Each red curve represents one selection from two optimization runs. Neither selected harness exceeds the Qwen3-32B baseline on the hard subset.}
\label{fig:hotpot-passk}
\par\medskip
\input{figures/latex/appworld.tex}
\caption{AppWorld \passat{k} curves for the GPT-OSS-120B baseline ($H_0$) and selected optimized harnesses ($H^\star$) at medium reasoning effort, with a high-effort baseline. Each red curve represents one selection from two optimization runs. Neither selected harness exceeds the high-effort baseline on the hard subset.}
\label{fig:appworld-passk}
\end{figure}

\begin{observationbox}
Aggregate curves can conceal shortfalls on hard tasks.
As shown in Figures~\ref{fig:hotpot-passk} (a) and~\ref{fig:appworld-passk} (a), the selected harnesses approach or exceed their stronger references when evaluated across all tasks.
However, as shown in Figures~\ref{fig:hotpot-passk} (d) and ~\ref{fig:appworld-passk} (d), these pooled curves obscure a consistent pattern on the hard subsets: none of the selected harnesses exceeds its stronger reference at any evaluated attempt budget.
Stratifying the curves by difficulty labels provided by benchmark makes this contrast explicit.
On easy HotpotQA questions, the selected harnesses with 8B model match or exceed the stronger 32B reference throughout the evaluated budgets (Figure~\ref{fig:hotpot-passk} (b)); 
on easy AppWorld tasks, all systems eventually reach complete observed coverage (Figure~\ref{fig:appworld-passk} (b)).
On hard HotpotQA questions, the selected harnesses improve \passat{1} over the 8B baseline but yield little additional coverage at \passat{128}, reaching $70.73$-$73.17\%$, compared with $70.73\%$ for the baseline and $78.05\%$ for the 32B reference  (Figure~\ref{fig:hotpot-passk} (d)).
On hard AppWorld tasks, coverage gains over the medium-effort baseline are larger: the selected harnesses reach $88.89$-$92.06\%$ at \passat{128}, compared with the baseline's $82.54\%$, but at best match the high-effort reference's $92.06\%$  (Figure~\ref{fig:appworld-passk} (d)).
Thus, viewing the curves only in aggregate would miss a consistent limitation: parity or gains over stronger references across all tasks do not translate into an advantage on hard tasks.
\end{observationbox}

\begin{figure}
\centering
\input{figures/latex/harness_gain.tex}
\caption{Task-level \passat{1} gains (pp) from harness optimization versus model scaling (Text: GPT-OSS-20B to 120B; HotpotQA: Qwen3-8B to 32B) or reasoning effort (AppWorld: medium to high). Small markers denote tasks; outlined symbols show equal-task group means. The diagonal indicates equal gains.
}
\label{fig:harness-gain}
\centering
\input{figures/latex/hotpotqa_evolution.tex}
\caption{HotpotQA \passat{k} curves for all 40 Qwen3-8B candidates generated in one optimization run, including the selected harness (red). On the hard subset, all candidate curves remain below the Qwen3-32B baseline.}
\label{fig:hotpot-evolution}
\vspace{-0.2in}
\end{figure}

\begin{figure}
\centering
\input{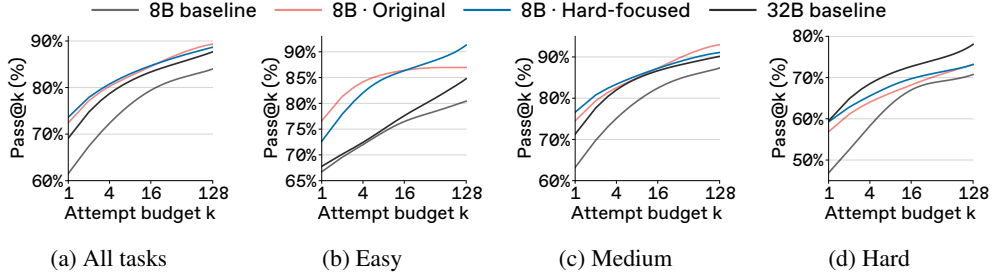}
\caption{\passat{k} curves for Qwen3-8B harnesses optimized with the original and hard-focused task mixtures, alongside the 8B and 32B baselines, overall and by difficulty.}
\label{fig:hotpotqa-difficulty-optimization}
\end{figure}

\begin{discussionbox}

\textbf{Across all three benchmark panels, scaling yields larger mean \passat{1} gains on hard groups than harness optimization.}
In Figure~\ref{fig:harness-gain}, we compare task-level \passat{1} gains from harness optimization and model or effort scaling relative to the same original baseline.
Here, the text classification groups are dataset-level proxies informed by the success distributions in Figure~\ref{fig:static-task-difficulty}, whereas HotpotQA and AppWorld use their original difficulty labels.
Across all three benchmark panels, the hard group means consistently favor scaling, highlighting that aggregate performance can conceal differences in where the two interventions provide their gains.

\textbf{The hard task shortfall is shared by all candidates during the evolution.}
We study all 40 Qwen3-8B candidate harnesses from one optimization run in Figure~\ref{fig:hotpot-evolution}.
On easy and medium questions, some candidates match or exceed the 32B reference, with their relative performance varying across attempt budgets.
On hard questions, however, every candidate's \passat{k} curve remains below the 32B reference throughout $k$.
Alternative candidates offer higher hard task coverage than the selected harness, but still fall short of the stronger reference.
The selected harness achieves the highest hard task \passat{1} among these candidates, yet does not attain the highest \passat{128}.

\textbf{The hard task shortfall persists even after optimization is redirected.}
One possible way to address the shared hard task shortfall is to give hard questions greater emphasis during optimization.
To this end, we re-optimize the Qwen3-8B harness with a hard focused task mixture and examine whether this targeted intervention broadens coverage on hard test questions.
As shown in Figure~\ref{fig:hotpotqa-difficulty-optimization} (d), the resulting harness improves hard task \passat{1} from the original optimized harness's $56.88\%$ to $59.28\%$, nearly matching the 32B reference's $59.51\%$.
At \passat{128}, however, coverage remains at $73.17\%$, unchanged from the original optimized harness and $4.88$ percentage points below the reference's $78.05\%$.
Thus, even after explicitly redirecting optimization toward hard questions, the gain amounts to more frequent success without an increase in the number of hard questions solved within 128 attempts.
\end{discussionbox}

\begin{figure}
\centering
\input{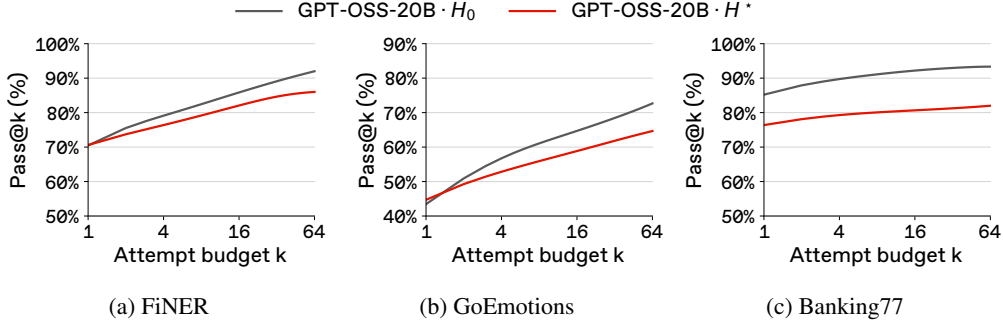}
\caption{
Harness transfer to new text-classification datasets.
\passat{k} curves for GPT-OSS-20B with the baseline ($H_0$) and in-distribution selected harness ($H^\star$) on out-of-distribution benchmarks; FiNER, GoEmotions, and Banking77.}
\label{fig:static-passk-ood}
\end{figure}

\begin{takeawaybox}
When hard task coverage matters, evaluate whether harness optimization expands coverage on that subset at the intended attempt budget.
Neither aggregate gains nor improved hard task \passat{1} after hard-focused optimization establish broader coverage.
Compare the optimized harness with feasible larger-model or higher-effort alternatives on the same subset.
\end{takeawaybox}

\subsection{Do Optimized Harnesses Transfer to OOD tasks?}

\begin{observationbox}
The distinction between \passat{1} and \passat{64} also matters when the selected harness is transferred to new domains. 
We transfer the ID-selected GPT-OSS-20B harness used in Section~\ref{sec:results-reference} to OOD domains: FiNER, GoEmotions, and Banking77, with both base and optimized harnesses given labeled target-domain examples. 
As shown in Figure~\ref{fig:static-passk-ood}, the transferred harness's \passat{64} falls below the baseline by $6.00$, $8.00$, and $11.33$ pp, respectively. 
Its small \passat{1} advantages on FiNER and GoEmotions already reverse at $k=2$.
\footnotemark
\end{observationbox}

\footnotetext{
Although random guessing can inflate coverage at larger attempt budgets~\citep{yue2025does}, both harnesses exceed uniform-guessing coverage on the four single-label datasets at smaller $k$, as detailed in Appendix~\ref{app:random-guessing}.
}

\begin{figure}
\input{figures/latex/text_classification_harness_code.tex}
\caption{Pseudocode derived from the base harness (a) and the subpart of the selected harness (b).}
\label{fig:text-harness-code}
\end{figure}

\begin{discussionbox}
To investigate possible mechanisms behind domain dependent harness behavior, we also review the selected harness.
Since optimization modifies explicit prompts and executable code while keeping solver parameters fixed, its decision rules are directly inspectable.
As illustrated in Figure~\ref{fig:text-harness-code}, 
the base harness directly returns the prediction from few-shot context, whereas the optimized harness selects prompt templates using the number and average length of labels observed in the supplied examples.
Its catalog template permits multiple labels separated by semicolons, a convention compatible with LawBench, where a case can involve multiple criminal charges.
However, the same routing heuristic selects this template for Banking77, which requires exactly one intent.
For the query ``I need a new card and I live in the United States.'', an OOD prediction was \texttt{order\_physical\_card;country\_support}, while the gold label was \texttt{country\_support}.
This explains the curve pattern for OOD tasks.

\end{discussionbox}
\begin{takeawaybox}
For OOD deployment, validate the transferred harness against the target-domain baseline.
Use code inspection to identify domain dependent assumptions for targeted testing.
\end{takeawaybox}

\section{Conclusion}
\label{sec:conclusion}
We provide a diagnostic lens for understanding what harness optimization improves, where its gains persist, and where performance shortfalls remain.
By combining \passat{k} curves with stronger reference systems and analyses across task difficulty, we distinguish more reliable success from broader observed task coverage.
These diagnostics help assess whether a harness suits an intended attempt budget and task group.

\textbf{Limitations and future work.}
Achieving consistent capability expansion on hard tasks remains an important open challenge for harness optimization, especially when successful executions support recursive self-improvement or evolutionary search.
This motivates search and selection methods that reward validated successes on previously uncovered hard tasks and retain candidates with complementary successes.
Large-budget \passat{k} analysis provides a diagnostic basis.
Our experimental results are bounded by the evaluated benchmarks, solver families, and optimization runs examined in this study.

\bibliographystyle{iclr2027_conference}
\bibliography{iclr2027_conference}

\appendix
\clearpage

\section{Repeated Evaluation and \texorpdfstring{\passat{k}}{pass@k}}
\label{app:passk}

We repeat complete executions of each fixed solver--harness pair on the same $N$ test tasks, including all internal model calls and tool interactions. Let $n$ be the number of rollouts and $c_i$ be the number of successes of task $x_i$. We estimate coverage within $k$ attempts following \citet{chen2021evaluating}:
\begin{equation}
\hat{\mathrm{pass@}k}(M,H)
=\frac{1}{N}\sum_{i=1}^{N}
\left[1-\frac{\binom{n-c_i}{k}}{\binom{n}{k}}\right],
\qquad 1\leq k\leq n,
\label{eq:passk-estimator}
\end{equation}
where the numerator is zero when $n-c_i<k$. Tasks are weighted equally within each dataset or benchmark; text macro averages weight datasets equally. At $k=1$, this is mean task success; at $k=n$ for equal-sized pools, it is the fraction solved at least once. Equation~\ref{eq:passk-estimator} is unbiased under independent, identically distributed attempts from a fixed policy.

\section{Experimental Details}
\label{app:protocols}

\subsection{Datasets and Splits}
\label{app:datasets}

We follow prior split protocols, with sample sizes and adaptations in Table~\ref{tab:appendix-datasets}. Symptom2Disease, LawBench, USPTO, and FiNER use the MCE splits~\citep{ye2026meta} distributed with Meta-Harness~\citep{lee2026metaharness}; GoEmotions~\citep{demszky2020goemotions} and Banking77~\citep{casanueva2020efficient} derive from their public releases. We subsample only examples with one annotated label from official splits for GoEmotions to ease the evaluation. HotpotQA follows GEPA's split sizes~\citep{agrawal2026gepa} using disjoint subsets of the official fullwiki training set. AppWorld retains the official train, development, and \path{test_normal} partitions~\citep{trivedi2024appworld}. Both dynamic benchmarks retain their original difficulty labels.
\begin{table}[htbp]
\centering
\footnotesize
\setlength{\tabcolsep}{4pt}
\renewcommand{\arraystretch}{1.1}
\caption{Dataset sizes and split roles. Prep. supplies examples or tasks to the harness; search/validation supplies optimization feedback. Reused selection shares the validation set.}
\label{tab:appendix-datasets}
\begin{tabular}{@{}lrrrr@{}}
\toprule
Dataset & Prep. & Search/val. & Select. & Test \\
\midrule
Symptom2Disease (ID) & 200 & 50 & reused & 212 \\
LawBench (ID) & 200 & 50 & reused & 100 \\
USPTO (ID) & 50 & 30 & reused & 100 \\
\addlinespace
FiNER (OOD) & 200 & --- & --- & 100 \\
GoEmotions (OOD) & 200 & --- & --- & 150 \\
Banking77 (OOD) & 200 & --- & --- & 150 \\
\addlinespace
HotpotQA & 150 & 300 & 300 & 300 \\
AppWorld & 30 & 60 & 57 & 168 \\
\bottomrule
\end{tabular}

\end{table}

\subsection{Baselines and Harness Optimization}
\label{app:harness-optimization}

We use Meta-Harness~\citep{lee2026metaharness} with Claude Opus 5 as the coding-agent proposer, keeping solver weights fixed. Each optimization allows 20 iterations with up to two proposals per iteration.

\paragraph{Text classification.}
We use GPT-OSS-20B for the baseline and optimized harness, with GPT-OSS-120B as the stronger reference; all use medium reasoning effort and temperature 0. Both baselines adapt the few-shot harness of \citet{lee2026metaharness}, resampling demonstrations for each query and repetition within a pre-defined character budget. We select by equal-dataset mean validation accuracy. The selected harness rebuilds memory from preparation examples for each repetition. OOD evaluation uses GPT-OSS-20B with the same preparation policy on target-domain examples, without further code optimization. 

\paragraph{HotpotQA.}
We use Qwen3-8B with Qwen3-32B as the stronger reference, both at temperature 0.6. The baseline follows GEPA's multihop seed program~\citep{agrawal2026gepa} with BM25 retrieval. 
The optimized harnesses are capped at four solver calls, two retrievals, and seven passages per retrieval as well as the baseline. We shortlist candidates using validation performance and select by exact match (EM) on a separate selection set.

\paragraph{AppWorld.}
We use GPT-OSS-120B at medium reasoning effort, with a high-effort baseline as the stronger reference. Repeated evaluation uses temperature 1. The baseline ports AppWorld's simplified ReAct agent~\citep{trivedi2024appworld}. Each attempt permits 50 solver calls and 50 environment steps, with a 2,400-second timeout extended to 7,200 seconds for high effort. Search and selection use temperature zero and original TGC, with final selection on the development set.
\subsection{Success Scorers}
\label{app:success-scorers}

We apply the same scorer to all systems within each benchmark, relaxing format requirements where necessary to better distinguish task completion from format compliance. In text classification, we use binary correctness after dataset-specific surface normalization. In HotpotQA, to accept correct answers embedded in longer responses, we use $\mathrm{Judge}=(\mathrm{Contain}\land\mathrm{CORRECT})$: answer containment with an LLM judgment confirming correctness.
Table~\ref{tab:hotpotqa-em-examples} gives examples that illustrate the failure mode from the Qwen3-32B baseline in Figure~\ref{fig:hotpot-passk}.
Each complete response states the reference answer but receives EM = 0. With 128 attempts per task, EM gives the Qwen3-32B lower \passat{1} than Qwen3-8B, which is counterintuitive, whereas Judge reverses this ordering.
In AppWorld, we use arity-fixed TGC: original successes remain successful, and an action attempt is additionally accepted only if all state checks pass, no execution error is recorded, and its sole failed check concerns an unrequested answer. With 128 attempts per task, original TGC ranks the GPT-OSS-120B below GPT-OSS-20B at \passat{1}; arity-fixed TGC reverses this ordering.\label{app:appworld-settings}
\begin{table}[htbp]
\caption{Exact-match false negatives from the Qwen3-32B HotpotQA baseline.
Bold spans mark reference answers in the complete recorded responses.}
\label{tab:hotpotqa-em-examples}
\vspace{4pt}
\centering
\begingroup
\small
\setlength{\tabcolsep}{3pt}
\renewcommand{\arraystretch}{1.15}
\hyphenpenalty=10000
\exhyphenpenalty=10000
\begin{tabular}{@{}
  >{\raggedright\arraybackslash}p{\dimexpr.44\linewidth-1.76\tabcolsep\relax}
  >{\raggedright\arraybackslash}p{\dimexpr.18\linewidth-.72\tabcolsep\relax}
  >{\raggedright\arraybackslash}p{\dimexpr.38\linewidth-1.52\tabcolsep\relax}@{}}
\toprule
Question & Reference answer & Complete recorded answer \\
\midrule
Which band was formed first, Loudspeakers or Marcy Playground? & Marcy Playground & \textbf{Marcy Playground} was formed first. \\
\addlinespace[6pt]
Which dog is larger, the Cirneco dell'Etna or the Siberian Husky? & Siberian Husky & The \textbf{Siberian Husky} is larger than the Cirneco dell'Etna. \\
\addlinespace[6pt]
John Alexander is the uncle of the player who plays what position for Liverpool? & right-back & The player is a \textbf{right-back} for Liverpool. \\
\addlinespace[6pt]
Intimidad de los parques was based on a short story by the novelist and essayist of what nationality? & Argentine & The novelist and essayist was \textbf{Argentine}. \\
\addlinespace[6pt]
When was the direct-to-video movie that borrows heavily from a 1988 American action film starring Bruce Willis released? & 1996 & The direct-to-video movie was released in \textbf{1996}. \\
\bottomrule
\end{tabular}
\endgroup
\end{table}

\clearpage
\section{Additional Experimental Results}
\label{app:additional-results}

\subsection{Variation Across Optimization Runs}
\label{app:run-variability}
\begin{figure}[!htbp]
\centering
\begingroup
\setlength{\parskip}{0pt}
\captionsetup[subfloat]{font=footnotesize,farskip=0pt,nearskip=0pt}
\setcounter{subfigure}{0}
\newcommand{\uncertaintyheading}[2]{%
  {\normalsize\rmfamily\mdseries
  \vphantom{#2}%
  \raisebox{-2.5pt}[0pt][0pt]{%
    \includegraphics{figures/uncertainty/#1-title.pdf}}\par}%
  \vspace{1.5pt}\nointerlineskip
}
\uncertaintyheading{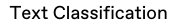}{Text Classification}
\includegraphics[width=\linewidth]{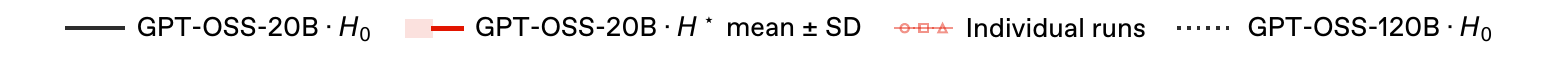}
\par\nointerlineskip
\subfloat[ID macro]{%
  \includegraphics[width=0.25\linewidth]{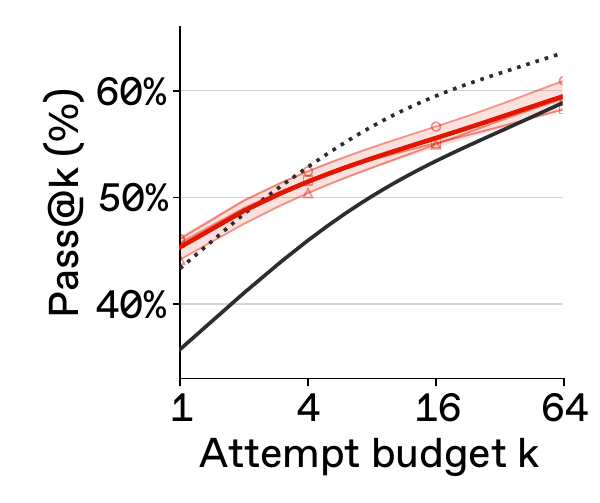}%
}%
\subfloat[Symptom2Disease]{%
  \includegraphics[width=0.25\linewidth]{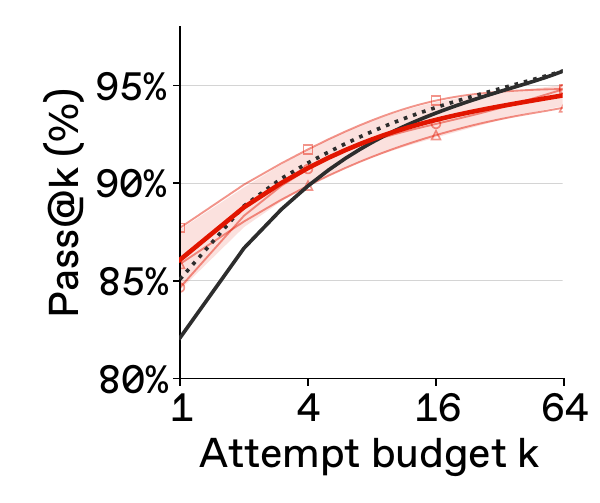}%
}%
\subfloat[LawBench]{%
  \includegraphics[width=0.25\linewidth]{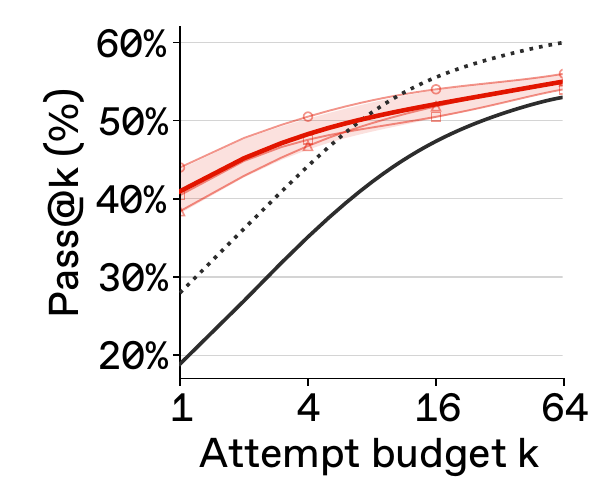}%
}%
\subfloat[USPTO]{%
  \includegraphics[width=0.25\linewidth]{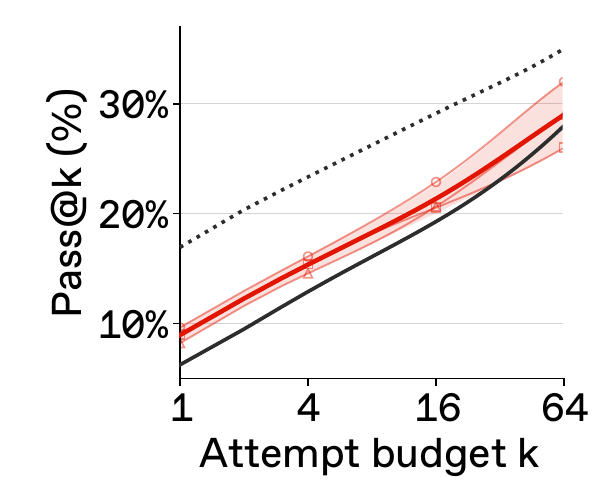}%
}%
\par
\vspace{10pt}
\uncertaintyheading{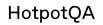}{HotpotQA}
\includegraphics[width=\linewidth]{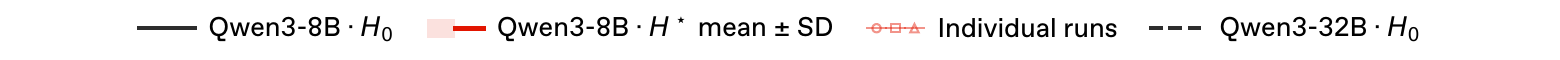}
\par\nointerlineskip
\subfloat[All tasks]{%
  \includegraphics[width=0.25\linewidth]{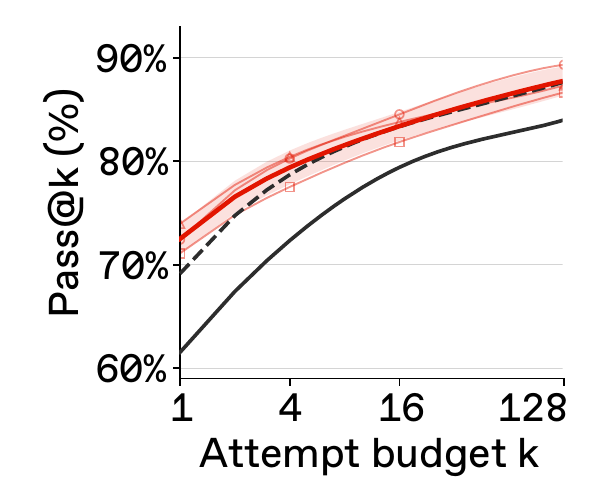}%
}%
\subfloat[Easy]{%
  \includegraphics[width=0.25\linewidth]{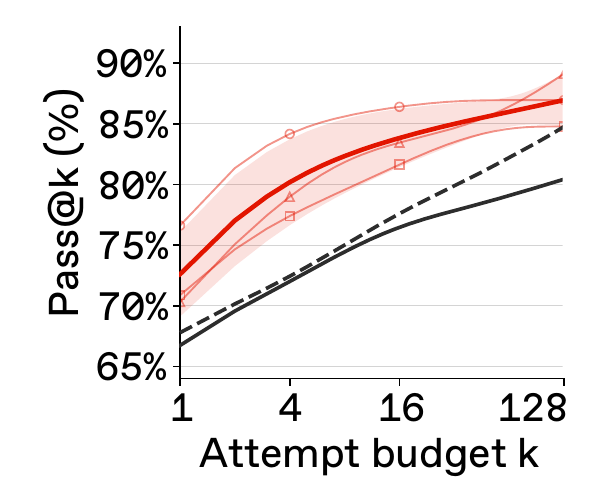}%
}%
\subfloat[Medium]{%
  \includegraphics[width=0.25\linewidth]{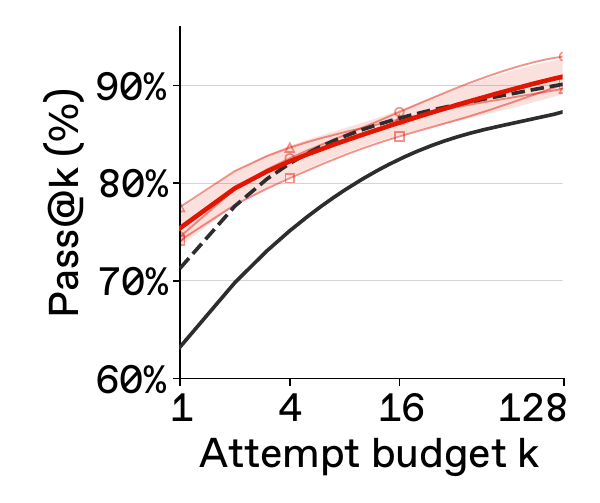}%
}%
\subfloat[Hard]{%
  \includegraphics[width=0.25\linewidth]{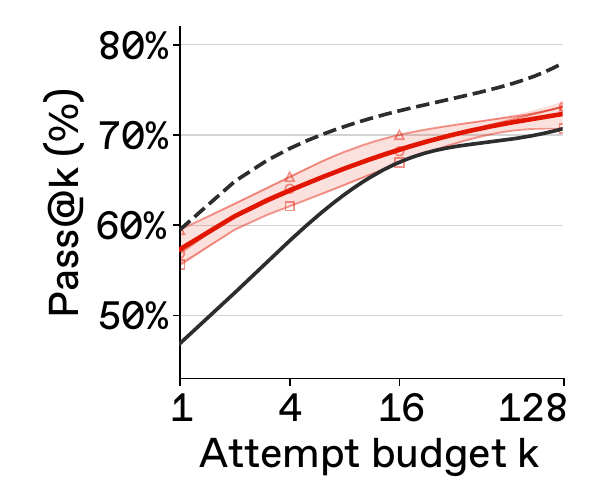}%
}%
\par
\vspace{10pt}
\uncertaintyheading{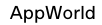}{AppWorld}
\includegraphics[width=\linewidth]{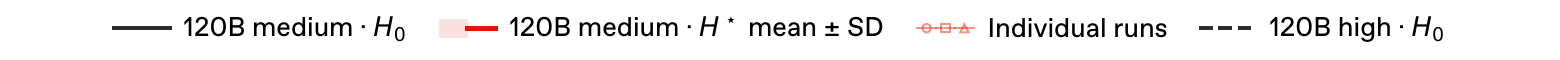}
\par\nointerlineskip
\subfloat[All tasks]{%
  \includegraphics[width=0.25\linewidth]{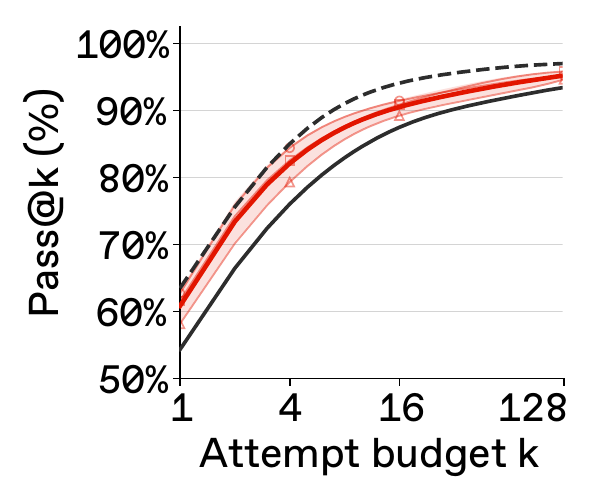}%
}%
\subfloat[Easy]{%
  \includegraphics[width=0.25\linewidth]{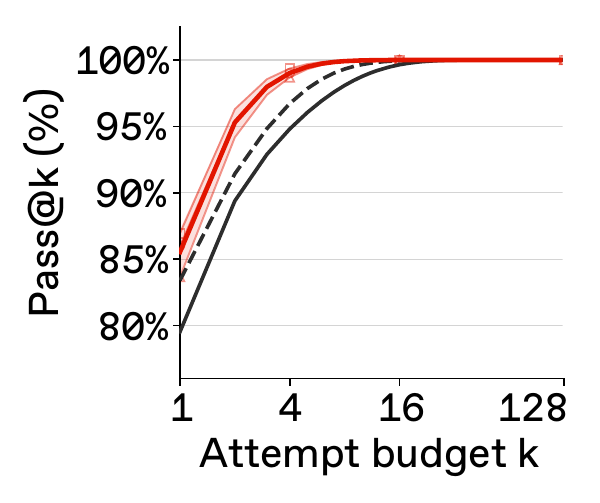}%
}%
\subfloat[Medium]{%
  \includegraphics[width=0.25\linewidth]{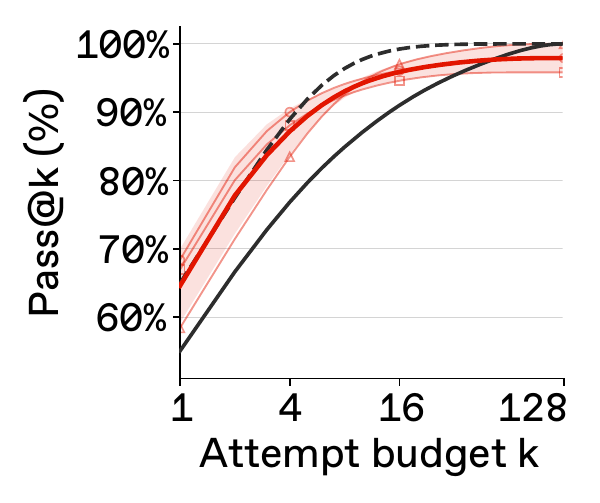}%
}%
\subfloat[Hard]{%
  \includegraphics[width=0.25\linewidth]{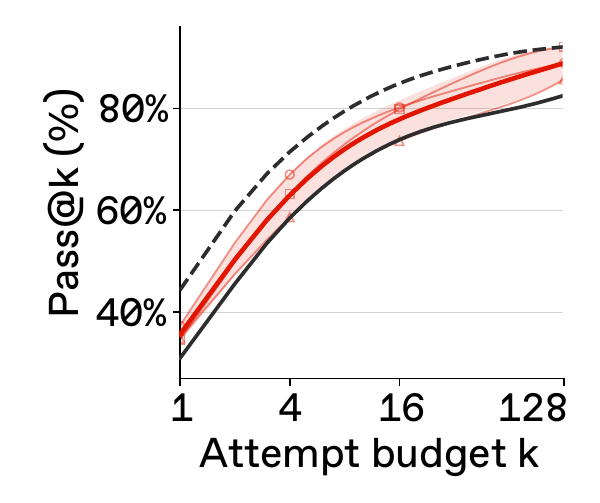}%
}%
\par
\caption{$\mathrm{pass@}k$ across three optimization runs:
mean $\pm1$ sample SD (thick red curves and bands), individual runs
(marked curves), and shared baselines (gray).}
\label{fig:uncertainty}
\endgroup
\end{figure}

We repeat harness optimization three times per benchmark and evaluate the selected harness from each run. Figure~\ref{fig:uncertainty} shows the mean and sample standard deviation (SD) across runs. Evaluation uses 64 attempts per task for text classification and 128 for HotpotQA and AppWorld.
Mean gains over the original baseline shrink from \passat{1} to the largest evaluated budget in all three benchmarks. The amount of the gain varies across runs, but improved single-attempt success does not consistently yield an advantage over stronger references at larger budgets. In text classification, all three optimized systems fall below the larger-model reference at \passat{64}. On hard HotpotQA and AppWorld tasks, none exceeds the larger-model or higher-effort reference, respectively, anywhere along the evaluated curves. 
SD includes evaluation and preparation; it does not isolate optimizer variance.

\clearpage

\subsection{Static Tasks}
\label{app:static-numbers}

\begin{figure}[htbp]
\centering
\input{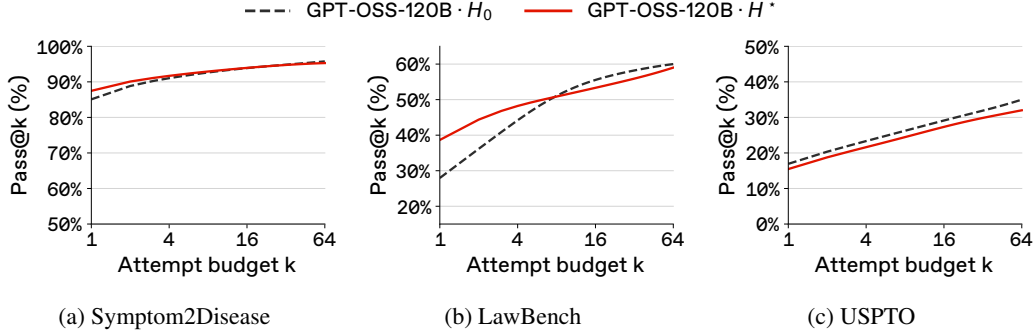}
\caption{
Text-classification \passat{k} curves for the GPT-OSS-120B baseline ($H_0$) and selected optimized harnesses ($H^\star$).}
\label{fig:text-classification-120b}
\end{figure}

\paragraph{Additional results on GPT-OSS-120B.}
Single-attempt gains also do not improve \passat{64} for GPT-OSS-120B. 
In Figure~\ref{fig:text-classification-120b}, LawBench \passat{1} increases from $27.95\%$ to $38.66\%$, while \passat{64} decreases from $60\%$ to $59\%$. 
The optimized harness also has lower \passat{64} on Symptom2Disease, and its USPTO curve remains below the baseline throughout.

\paragraph{Task-level changes in observed success sets.}
\label{app:success-decomposition}
Net coverage changes on the static tasks combine newly solved and lost tasks across each dataset, as shown in Table~\ref{tab:static-success-decomposition}.

\begin{table}[htbp]
\caption{Changes in observed success sets from the GPT-OSS-20B base harness ($H_0$) to the selected harness ($H^\star$) in Figure~\ref{fig:task-comparison} (a)-(c), with $n=k=64$. $S_0$ and $S_\star$ are the tasks with at least one success under $H_0$ and $H^\star$.}
\label{tab:static-success-decomposition}
\centering
\small
\setlength{\tabcolsep}{4pt}
\begin{tabular}{lrrrrrrrr}
\toprule
& & \multicolumn{3}{c}{Tasks} & \multicolumn{4}{c}{\passat{1} gain (pp)} \\
\cmidrule(lr){3-5}\cmidrule(lr){6-9}
Dataset & $|S_0|\!\to\!|S_\star|$ & $S_\star\setminus S_0$ & $S_0\setminus S_\star$ & $S_0\cap S_\star$ & Total & $S_0\cap S_\star$ & $S_\star\setminus S_0$ & $S_0\setminus S_\star$ \\
\midrule
Symptom2Disease & $203\to201$ & 3 & 5 & 198 & $+2.57$ & $+2.63$ & $+0.38$ & $-0.43$ \\
LawBench & $53\to56$ & 7 & 4 & 49 & $+25.14$ & $+21.02$ & $+4.64$ & $-0.52$ \\
USPTO & $28\to32$ & 7 & 3 & 25 & $+3.39$ & $+3.25$ & $+0.19$ & $-0.05$ \\
\bottomrule
\end{tabular}
\end{table}

\paragraph{Frozen memory.}
\label{app:frozen-memory}
Reusing one prepared memory across repetitions, instead of rebuilding it for each repetition, leaves the pattern in Figure~\ref{fig:task-comparison} (a)-(c) unchanged, as shown in Table~\ref{tab:static-frozen-memory}.
With either memory policy, the selected GPT-OSS-20B harness improves \passat{1} over $H_0$ on all three datasets, while at \passat{64} it falls below $H_0$ on Symptom2Disease, its LawBench gain narrows, and it remains below the GPT-OSS-120B reference on every dataset.

\begin{table}[htbp]
\caption{\passat{1} and \passat{64} (\%) for the text-classification in Figure~\ref{fig:task-comparison} (a)-(c), with $n=64$. Rebuilt and Frozen denote the selected GPT-OSS-20B harness ($H^\star$) with memory rebuilt for each repetition, as in the figure, or prepared once per dataset and reused.}
\label{tab:static-frozen-memory}
\centering
\small
\setlength{\tabcolsep}{4pt}
\begin{tabular}{lrrrrrrrr}
\toprule
& \multicolumn{4}{c}{\passat{1}} & \multicolumn{4}{c}{\passat{64}} \\
\cmidrule(lr){2-5}\cmidrule(lr){6-9}
Dataset & 20B $H_0$ & Rebuilt & Frozen & 120B $H_0$ & 20B $H_0$ & Rebuilt & Frozen & 120B $H_0$ \\
\midrule
Symptom2Disease & 82.08 & 84.66 & 86.34 & 85.08 & 95.75 & 94.81 & 92.45 & 95.75 \\
LawBench & 18.86 & 44.00 & 43.53 & 27.95 & 53.00 & 56.00 & 53.00 & 60.00 \\
USPTO & 6.25 & 9.64 & 10.03 & 16.94 & 28.00 & 32.00 & 30.00 & 35.00 \\
\bottomrule
\end{tabular}
\end{table}

\clearpage
\paragraph{Random guessing on static tasks.}
\label{app:random-guessing}
Both harnesses exceed uniform-guessing coverage on the four single-label datasets at smaller $k$, as shown in Table~\ref{tab:static-guessing-comparison}. Independent uniform guesses over $L$ labels yield expected coverage $1-(1-1/L)^k$.

\begin{table}[hbt]
\caption{Coverage (\%) for uniform-label guessing and the GPT-OSS-20B baseline ($H_0$) and selected harness ($H^\star$) in Figures~\ref{fig:task-comparison} and~\ref{fig:static-passk-ood}. Guessing values are analytical; harness estimates use $n=64$.}
\label{tab:static-guessing-comparison}
\centering
\small
\setlength{\tabcolsep}{5pt}
\begin{tabular}{lrrrrrr}
\toprule
& \multicolumn{3}{c}{\passat{4}} & \multicolumn{3}{c}{\passat{8}} \\
\cmidrule(lr){2-4}\cmidrule(lr){5-7}
Dataset & Guessing & $H_0$ & $H^\star$ & Guessing & $H_0$ & $H^\star$ \\
\midrule
\multicolumn{7}{l}{\textit{In-distribution}} \\
Symptom2Disease & 16.98 & 89.86 & 90.72 & 31.08 & 92.08 & 92.16 \\
LawBench & --- & 35.12 & 50.51 & --- & 42.15 & 52.59 \\
USPTO & --- & 12.90 & 16.10 & --- & 16.08 & 19.30 \\
\midrule
\multicolumn{7}{l}{\textit{Out-of-distribution}} \\
FiNER & 29.39 & 79.11 & 76.39 & 50.15 & 82.44 & 79.17 \\
GoEmotions & 13.54 & 56.74 & 52.84 & 25.24 & 61.08 & 55.92 \\
Banking77 & 5.09 & 89.73 & 79.29 & 9.93 & 91.10 & 80.07 \\
\bottomrule
\end{tabular}
\par\smallskip
\begin{minipage}{\linewidth}
\footnotesize
Guessing samples the complete label inventory uniformly with replacement. This single-label reference is not applicable to LawBench and USPTO, which require open-ended structured outputs.
\end{minipage}
\end{table}

\subsection{Dynamic Tasks}
\label{app:dynamic-results}

\begin{figure}[htbp]
\centering
\input{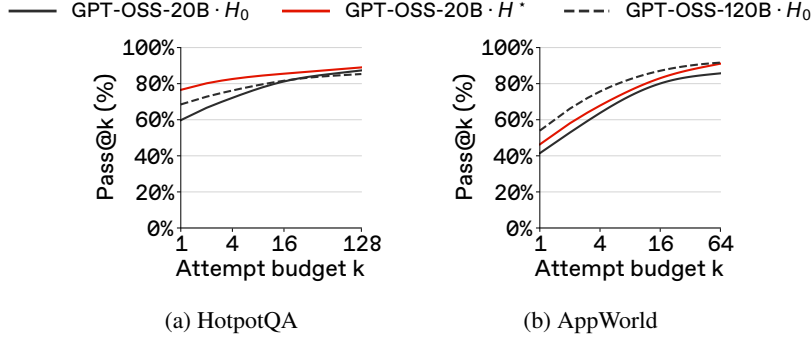}
\caption{
\passat{k} curves for the GPT-OSS-20B baseline ($H_0$) and selected optimized harnesses ($H^\star$), with a GPT-OSS-120B baseline.}
\label{fig:dynamic-gpt20b}
\label{fig:hotpotqa-gpt20b}
\label{fig:appworld-passk-20b}
\end{figure}

We also optimize a harness for GPT-OSS-20B on each dynamic benchmark, as shown in Figure~\ref{fig:dynamic-gpt20b}. The optimized system exceeds the 120B baseline on HotpotQA at \passat{128} ($89.00\%$ versus $85.33\%$), while remaining slightly below it on AppWorld at \passat{64} ($91.07\%$ versus $91.67\%$).

\begin{figure}[!htbp]
\centering
\begingroup
\setlength{\parskip}{0pt}
\captionsetup[subfloat]{farskip=0pt}
\setcounter{subfigure}{0}
\includegraphics[width=\linewidth]{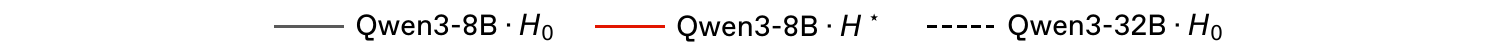}
\par\nointerlineskip
\subfloat[All tasks]{%
  \includegraphics[width=0.24\linewidth]{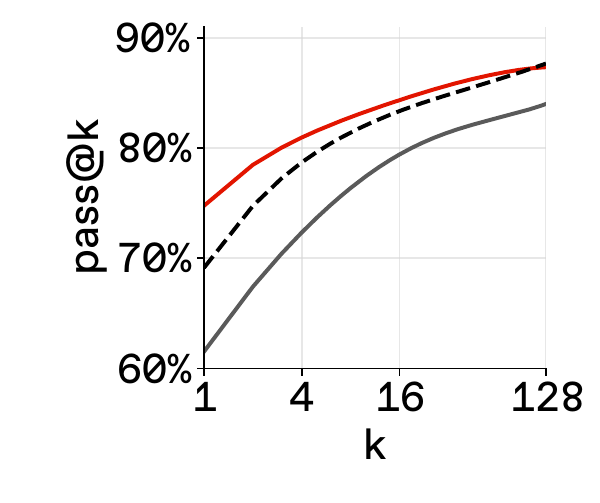}%
}
\subfloat[Easy]{%
  \includegraphics[width=0.24\linewidth]{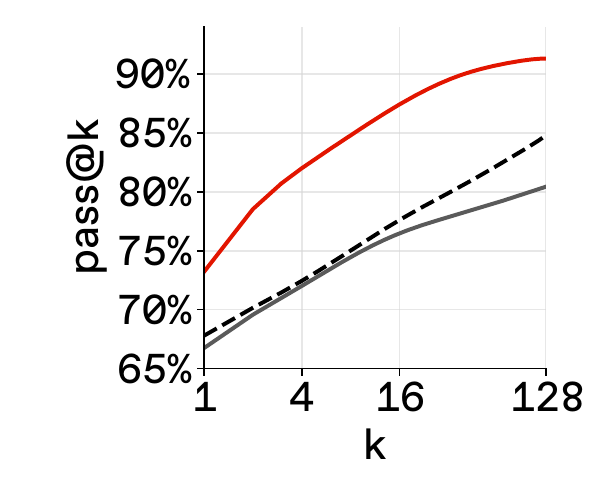}%
}
\subfloat[Medium]{%
  \includegraphics[width=0.24\linewidth]{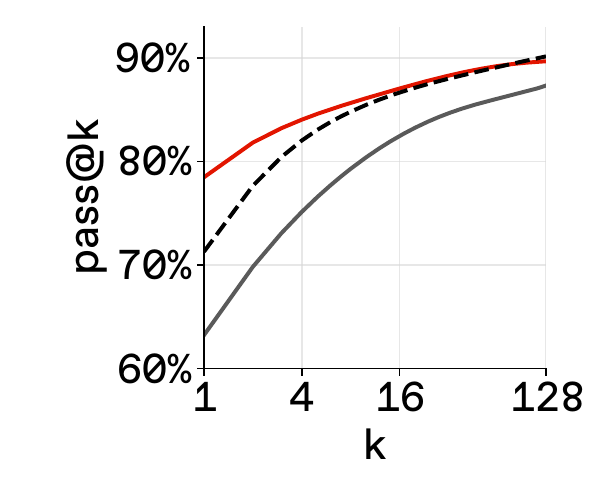}%
}
\subfloat[Hard]{%
  \includegraphics[width=0.24\linewidth]{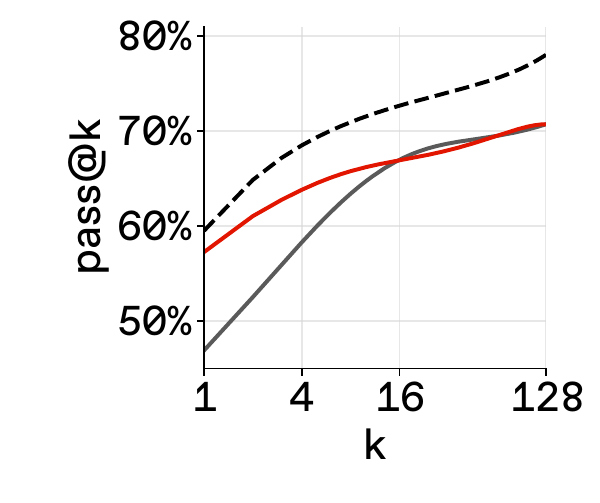}%
}
\par
\endgroup
\caption{HotpotQA $\mathrm{pass@}k$ with Judge used for optimization, selection, and evaluation.}
\label{fig:hotpotqa-scorer}
\end{figure}

\begin{figure}[!htbp]
\centering
\begingroup
\setlength{\parskip}{0pt}
\captionsetup[subfloat]{farskip=0pt}
\setcounter{subfigure}{0}
\includegraphics[width=\linewidth]{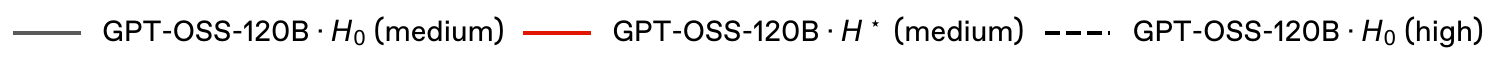}
\par\nointerlineskip
\subfloat[All tasks]{%
  \includegraphics[width=0.24\linewidth]{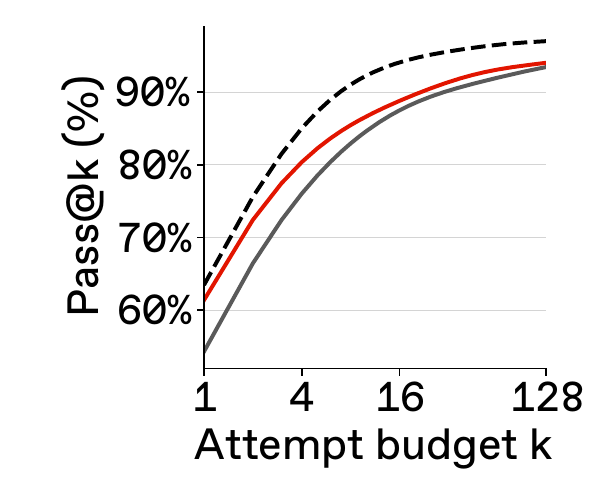}%
}
\subfloat[Easy]{%
  \includegraphics[width=0.24\linewidth]{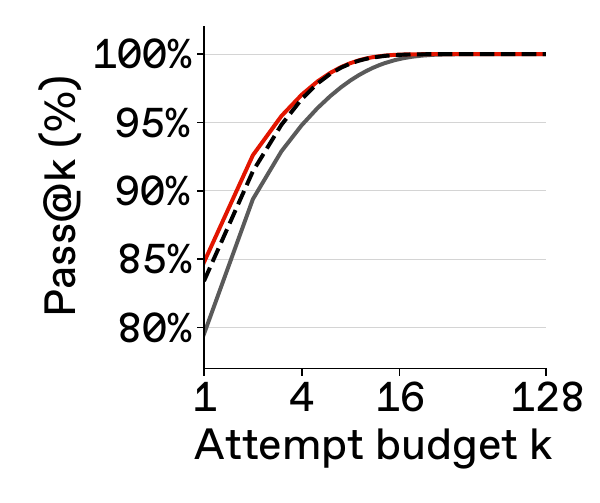}%
}
\subfloat[Medium]{%
  \includegraphics[width=0.24\linewidth]{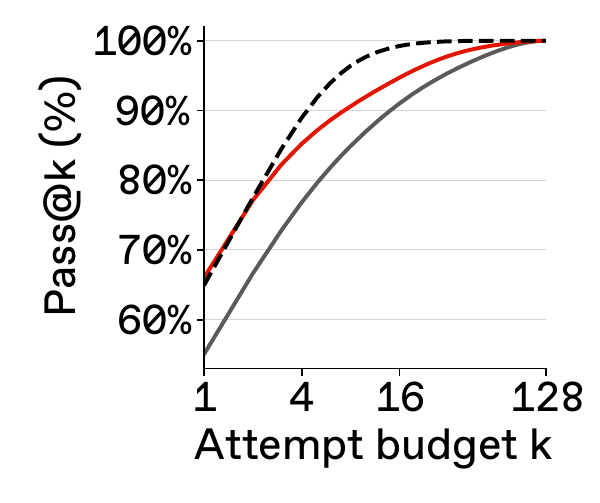}%
}
\subfloat[Hard]{%
  \includegraphics[width=0.24\linewidth]{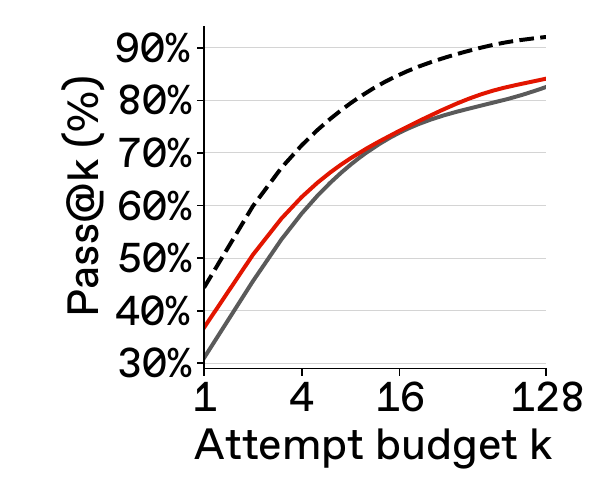}%
}
\par
\endgroup
\caption{AppWorld $\mathrm{pass@}k$ with arity-fixed TGC used for optimization, selection, and evaluation.}
\label{fig:appworld-scorer}
\end{figure}

\clearpage
\paragraph{Optimization with aligned scorers.}
\label{app:aligned-scorers}
We repeat optimization and selection with the success scorers defined in Appendix~\ref{app:success-scorers}. We retain the original test sets and difficulty groups and evaluate through $k=128$. Figures~\ref{fig:hotpotqa-scorer} and~\ref{fig:appworld-scorer} are consistent with the main findings in Section~\ref{sec:results-difficulty}. Overall gains over the original baselines narrow from \passat{1} to \passat{128}, while the displayed harnesses remain below the larger-model or higher-effort reference on hard tasks throughout $k$.

\subsection{Optimization Conditions}
\label{app:optimization-conditions}
\label{app:hotpot-resources}

Relaxed resource limits do not yield higher mean coverage at $k=128$ in these runs. We compare three Qwen3-8B optimization runs per setting, with caps of 4 solver calls, 2 retrievals, and 7 passages per retrieval (tight) versus 12/12/20 (relaxed). Each run proposes 40 candidates on the same optimization splits; the caps apply during both optimization and evaluation.

\begin{table}[htbp]
\centering
\small
\setlength{\tabcolsep}{2pt}
\caption{HotpotQA Judge scores (\%) on 300 test questions with $n=128$, reported as mean $\pm$ sample SD across three runs per cap setting.}
\label{tab:hotpot-optimization-budgets}
\begin{tabular*}{0.5\linewidth}{@{\extracolsep{\fill}}lrr@{}}
\toprule
Caps & \passat{1} & \passat{128} \\
\midrule
Relaxed & $73.39 \pm 1.60$ & $87.22 \pm 1.54$ \\
Tight & $72.50 \pm 1.43$ & $87.78 \pm 1.39$ \\
\bottomrule
\end{tabular*}

\end{table}

\end{document}